\documentclass[letterpaper]{article} 
\usepackage{aaai2026}  
\usepackage{times}  
\usepackage{helvet}  
\usepackage{courier}  
\usepackage[hyphens]{url}  
\usepackage{graphicx} 
\usepackage{natbib}  
\usepackage{caption} 
\usepackage{algorithm}
\usepackage{algorithmic}
\usepackage{amsmath,amssymb}  
\usepackage{bm}  
\usepackage{multirow}
\usepackage{comment}
\usepackage{newfloat}
\usepackage{listings}
\usepackage{booktabs}
\usepackage{pdfpages}
\usepackage{enumitem,pifont,tabularx}
\newcommand{\y}{\textbf{Yes}}

\DeclareCaptionStyle{ruled}{labelfont=normalfont,labelsep=colon,strut=off} 
\floatstyle{ruled}
\newfloat{listing}{tb}{lst}{}
\floatname{listing}{Listing}
\title{$\Delta t$-Mamba3D: A Time‑Aware Spatio‑Temporal \\State‑Space Model for Breast Cancer Risk Prediction}
\author {
    Zhengbo Zhou\textsuperscript{\rm 1},
    Dooman Arefan\textsuperscript{\rm 2},
    Margarita Zuley\textsuperscript{\rm 2},
    Shandong Wu\textsuperscript{\rm 1,2,3}
}
\affiliations {
    \textsuperscript{\rm 1}Intelligent Systems Program\\
    \textsuperscript{\rm 2}Department of Radiology\\
    \textsuperscript{\rm 3}Department of Biomedical Informatics and Department of Bioengineering
    \\University of Pittsburgh , Pittsburgh, PA, USA 
    
}

\usepackage{bibentry}

\begin{document}
\nocopyright
\maketitle

\begin{abstract}
Longitudinal analysis of sequential radiological images is hampered by a fundamental data challenge: how to effectively model a sequence of high-resolution images captured at irregular time intervals. This data structure contains indispensable spatial and temporal cues that current methods fail to fully exploit. Models often compromise by either collapsing spatial information into vectors or applying spatio-temporal models that are computationally inefficient and incompatible with non-uniform time steps. We address this challenge with Time-Aware $\Delta t$-Mamba3D, a novel state-space architecture adapted for longitudinal medical imaging. Our model simultaneously encodes irregular inter-visit intervals and rich spatio-temporal context while remaining computationally efficient. Its core innovation is a continuous-time selective scanning mechanism that explicitly integrates the true time difference between exams into its state transitions. This is complemented by a multi-scale 3D neighborhood fusion module that robustly captures spatio-temporal relationships. In a comprehensive breast cancer risk prediction benchmark using sequential screening mammogram exams, our model shows superior performance, improving the validation c-index by 2–5 percentage points and achieving higher 1–5 year AUC scores compared to established variants of recurrent, transformer, and state-space models. Thanks to its linear complexity, the model can efficiently process long and complex patient screening histories of mammograms, forming a new framework for longitudinal image analysis.

\end{abstract}


\section{Introduction}

Screening mammography for breast cancer detection is inherently longitudinal. Women return every a few years, image acquisition protocols evolve over time, breasts change with aging, and subtle preclinical lesions may emerge gradually across exams. Radiologists routinely examine longitudinal and cross-view information: they compare current and prior exams and craniocaudal (CC) and mediolateral oblique (MLO) views, assess side-to-side asymmetries, and evaluate interval change when estimating malignancy risk or assigning BI-RADS diagnostic categories \citep{scutt2006breast}.
Yet most deep learning systems for breast imaging still operate on a single imaging exam, ignoring the temporal context that drives clinical decision making \citep{yala2021toward}.  When multiple exams are available, most methods first collapse each exam into a single per-visit feature vector and then apply a temporal model (e.g., RNNs or GRUs), thereby sacrificing fine-grained lesion morphology and growth patterns \citep{dadsetan2022deep}. In addition, the irregular time gaps between exams—an important predictor of breast cancer risk—are usually left unencoded by existing deep-learning approaches \citep{zhou2025longmambattn}.

Irregularly timed data frequently occurs in clinical settings, reflecting varying degrees of disease severity—patients with severe conditions tend to have more frequent hospital visits. Despite this, many existing methods, such as standard Recurrent Neural Networks (RNNs) and Transformers, treat patient visits as tokens placed on an evenly spaced temporal grid \cite{karaman2024longitudinal}, which discard valuable interval information. Over the past decade, several specialized models have been proposed to address irregular sampling in clinical time series. Time-discretized models, such as GRU-D \citep{che2018recurrent} and Time-aware LSTM \citep{nguyen2020time}, incorporate elapsed time  or its exponential decay directly into hidden state updates. Although these methods capture interval magnitude efficiently, they inherently assume piecewise constant dynamics, limiting their ability to model evolving risk between observations. Continuous-time approaches, including Neural ODEs and Neural CDEs \citep{rubanova2019latent,kidger2020neural}, and recent advancements like ContiFormer \citep{chen2023contiformer}, naturally handle irregular intervals and offer continuous-time predictions. However, they have primarily been evaluated on minute-level ECG or sensor data or low-dimensional EHR records. In contrast, screening intervals in medical imaging often span from 0.5 to 3 years, involving extremely high-dimensional features. Such imaging visits are inherently sparse, with each patient encounter represented as a discrete event at a specific integer timestamp, accompanied by a zero-valued signal in between, a scenario inadequately modeled by previous methods. As highlighted by recent advances such as Mamba \citep{gu2023mamba}, state-space models (SSMs) with adaptive, context-aware parameters offer enhanced capability for capturing long-range dependencies in dynamic systems. Although Mamba has significantly surpassed conventional recurrent models in domains such as language modeling, it has not explicitly encoded irregular time intervals. Consequently, adapting models like Mamba to effectively handle irregularly timed, high-dimensional imaging data remains largely unexplored and presents an urgent need within the field.

Capturing joint spatial–temporal patterns in longitudinal medical imaging data poses significant methodological and computational challenges. Early 3D CNNs, such as C3D \citep{tran2015learning}, I3D \citep{carreira2017quo}, apply cubic convolutions to densely sampled frame stacks; their computation scales cubically with spatial resolution and their receptive field remains inherently limited. Video vision transformers—exemplified by TimeSformer \citep{bertasius2021space}, ViViT \citep{arnab2021vivit}, and Video Swin \citep{liu2022video}—simultaneously encode spatial structure and densely, uniformly sampled temporal dynamics, yielding joint spatio-temporal representations. Full self-attention scales quadratically with the number of spatio-temporal tokens and becomes impractical when modeling longer longitudinal visits, a limitation that is magnified by the typically small sample sizes in clinical datasets. Furthermore, standard video vision transformers assume equal temporal spacing and do not reflect real-world inter-visit intervals. More recently, structured state-space sequence models (SSMs)—S4 and Mamba \citep{gu2021efficiently,gu2023mamba} and vision variants that devise 2-D/3-D scan orders or local fusion windows \citep{zhu2024vision,liu2024vmamba,xiao2024spatial}—achieve linear-time scanning, but they likewise assume uniformly spaced tokens and must stack multiple costly passes to absorb full 3-D context.

To bridge these gaps, we propose Time‑Aware $\Delta t$‑Mamba3D, a spatio‑temporal state‑space block with two defining features: i) $\Delta t$‑aware transitions: We find that driving the SSM solely with raw inputs as control signals is ill‑suited to irregularly sampled clinical data, leaving the model largely insensitive to the true time gaps. Instead, we modulate every selective-scan update with the true inter‑visit interval $\Delta t$, enabling continuous‑time memory decay or accumulation under irregular sampling while preserving the original content‑aware step size;
    ii) Multi‑scale depth‑wise 3D fusion: neighborhood-adaptive convolutions jointly encode spatial and temporal context at low cost. Our pipeline first converts each exam into a token sequence via a unidirectional sweep (Fig. 1).  
The token states then evolve through the closed‑form SSM transition whose step size is modulated by the true inter‑visit gap, thereby embedding irregular temporal information. These states are then refined by structure‑aware 3D convolutions that re‑weight neighbouring voxels across multiple receptive fields, and an observation head maps the fused state to output variables.
This novel design preserves lesion morphology, accommodates irregular timing, and scales linearly in memory with sequence length.
We embed Structure‑Aware~$\Delta t$‑Mamba 3D in a prediction pipeline that ingests up to eight prior screening mammogram exams (four views including CC and MLO views of left and right breasts per exam) and outputs year-specific hazards. On two mammogram datasets with irregular inter-exam intervals, our model outperforms time-aware models and spatio-temporal models, improving c-index and 1–5-year AUCs while maintaining linear memory growth. Our contributions are summarized as follows:
\begin{enumerate}
    \item We extend Mamba with a selective scanning mechanism whose state transition explicitly incorporates the true inter‑visit interval~$\Delta t$ at the image level. By utilizing irregular time spans, combined with inputs as control signals for SSM to achieve superior model performance.
    \item We embed a multiscale, depth‑wise 3D convolution block within the Mamba module to efficiently capture joint spatio‑temporal context.
    \item On two longitudinal mammography datasets with varying temporal patterns and different class distributions, our model surpasses recurrent, transformer, and visual‑SSM baselines in accuracy while maintaining linear memory and computing scaling.
\end{enumerate}

\begin{figure*}[t]
  \centering
  \includegraphics[width=0.8\linewidth]{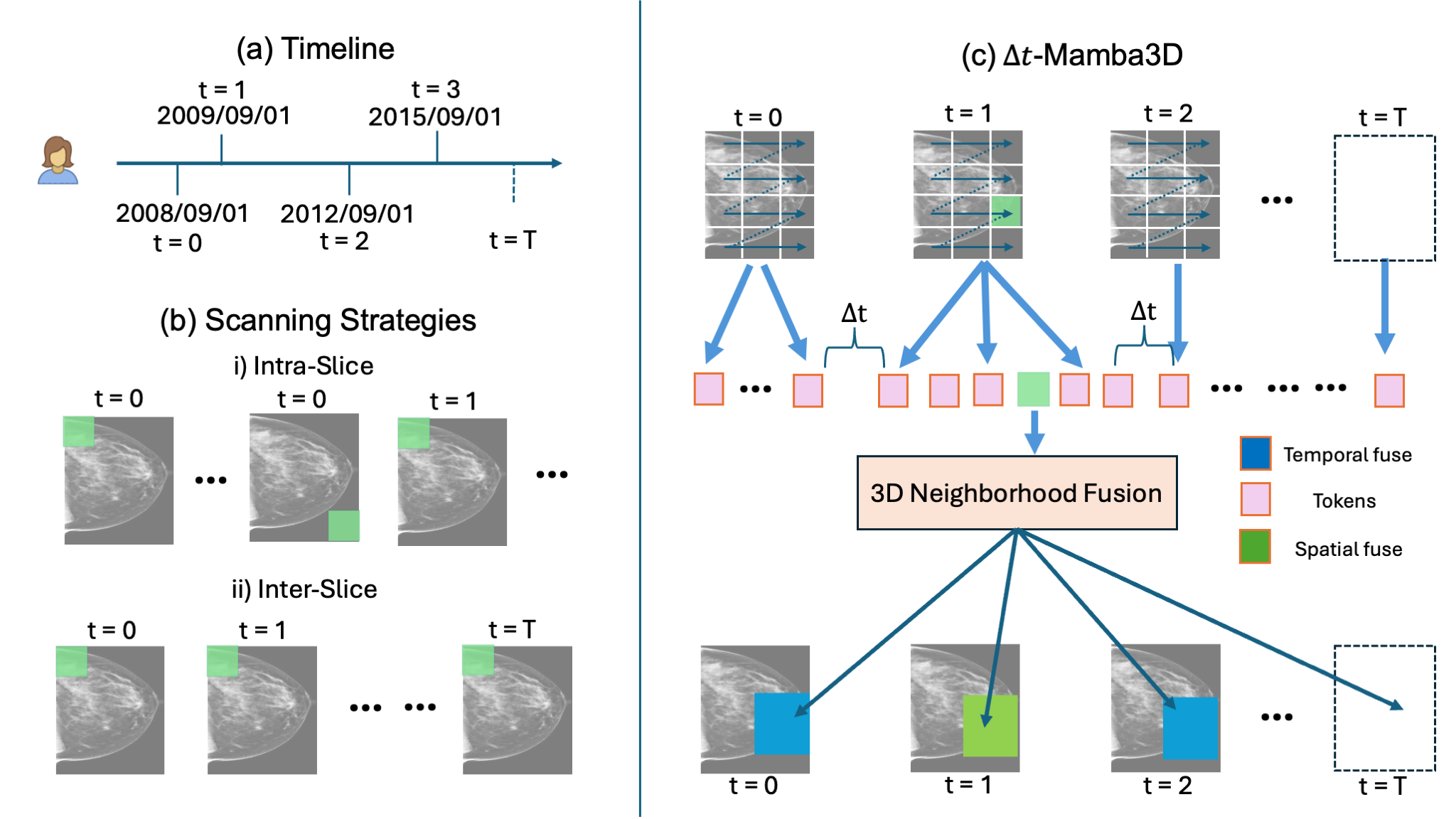}
  \caption{(a) Illustration of a patient's sequential imaging data acquired with irregular inter-visit gaps $\Delta t$ (e.g., 2009 → 2012 → 2015).
  (b) Different scanning strategies for spatio-temporal feature volumes. (c) The scanning mechanism in the proposed method $\Delta t$-Mamba3D: time-aware scan
  modulated by inter-visit gaps $\Delta t$ with learnable multi-scale
  3D neighborhood fusion.}
  \label{fig:sa_dt_mamba3d_scan}
\end{figure*}
\section{Related Work}

\subsection{Time-Aware Models}

Early models incorporated elapsed time between observations by explicitly modulating recurrent network updates. T-LSTM~\citep{nguyen2020time} and GRU-D~\citep{che2018recurrent} adopted data-driven exponential decay mechanisms to inputs and hidden states, effectively reducing the influence of outdated measurements. Attention-based methods have also integrated temporal information through various strategies. Approaches like time2vec~\citep{kazemi2019time2vec} introduced learnable temporal embeddings, while continuous-time attention models explicitly factor in the elapsed time  as positional biases or embedding components~\citep{shukla2021multi}. ContiFormer~\citep{chen2023contiformer}, a recent advancement, further enhances this line by leveraging continuous-time self-attention mechanisms specifically designed to handle irregularly sampled time series. Beyond these parametric decay and attention-based strategies, continuous-time latent dynamics have been extensively modeled by methods such as Latent ODEs~\citep{rubanova2019latent} and Neural Controlled Differential Equations (Neural CDEs)~\citep{kidger2020neural}. These approaches explicitly integrate hidden state trajectories between irregular event occurrences, effectively capturing complex continuous-time dependencies within data.

\subsection{Vision State Space Models}

Recent visual SSMs tackle spatial coherence by crafting tailored scan patterns \cite{liu2024vmamba,zhu2024vision}. Spatial-Mamba \citep{xiao2024spatial} tiles  2D patches and applies a local fusion window to mitigate scan-order bias. For 3D volumes and long-horizon video, Seg-Mamba \citep{xing2024segmamba} and LongMamba \citep{zhou2025longmambattn} tokenize slices or frames into patch sequences with bespoke spatial–temporal scan orders, ensuring that both intra-slice structure and inter-frame dynamics are captured.

\begin{figure}[t]
  \centering
  \includegraphics[width=\linewidth]{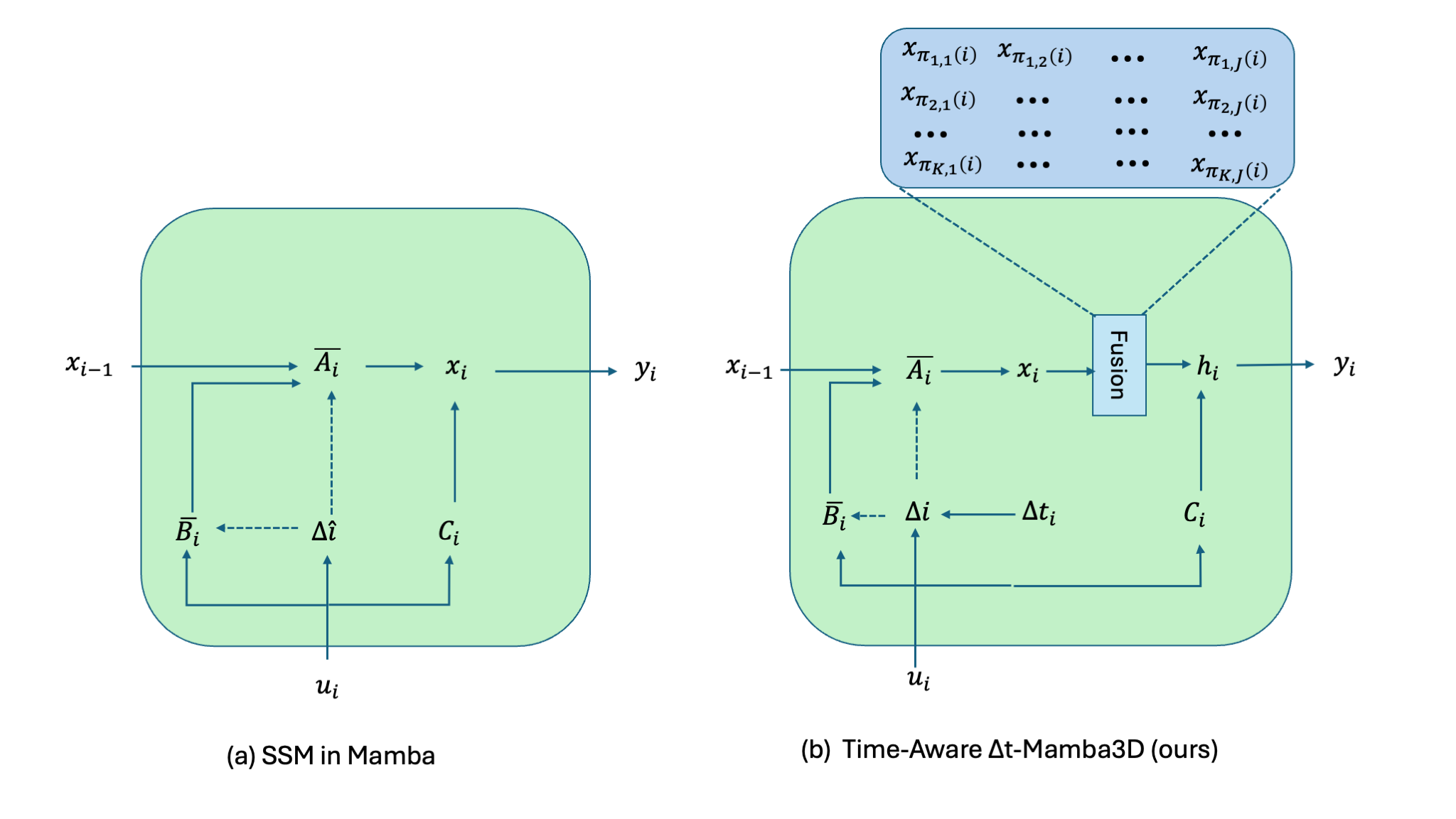}
  \caption{State-space modules. (a) Standard Mamba SSM: input $u_i$ produces
  parameters $(\bar A_i,\bar B_i,\delta_i,C_i)$ that update state $x_i$ and emit $y_i$.
   (b) Time-Aware
  $\Delta t$-Mamba3D (ours) generalizes to irregularly sampled spatio--temporal
  grids: each token's step size is modulated by the true inter-visit gap
  $\Delta t_i$, and a learnable multi-scale 3D neighborhood fusion aggregates local structure across visits before producing $h_i$ and $y_i$.}
  \label{fig:sa_dt_mamba3d_overall}
\end{figure}
\section{Methods}



\subsection{Preliminary}

SSMs are commonly used for analyzing sequential data and modeling continuous linear time-invariant systems \cite{williams2007linear}. This dynamic system can be described by the linear state transition and observation equations \cite{kalman1960new}. A standard linear continuous-time state-space model (SSM) can be expressed as:
\begin{equation}
\mathbf{h'}(i)=\mathbf{A}\mathbf{h}(i)+\mathbf{B}\mathbf{x}(i),\quad \mathbf{y}(i)=\mathbf{C}\mathbf{h}(i)+\mathbf{D}\mathbf{x}(i).
\label{eq:cts}
\end{equation}
where $\mathbf{A}$, $\mathbf{B}$ and $\mathbf{C}$ are the weighting trainable parameters, and $\mathbf{D}$ always equals to 0.
To effectively integrate continuous-time SSMs into the deep learning framework, it is essential to discretize the continuous-time models. Sampling this continuous-time SSM at intervals of size $\delta$ (assuming zero-order hold) yields the discrete counterpart:
\begin{equation}
\mathbf{h_i}=\bar{\mathbf{A}}\mathbf{h}_{i-1}+\bar{\mathbf{B}}\mathbf{x}_i,\quad \mathbf{y}_i=\mathbf{C}\mathbf{h}_i+\mathbf{D}\mathbf{x}_i,
\label{eq:dssm}
\end{equation}
with
\[
\bar{\mathbf{A}} = e^{\delta \mathbf{A}}, 
\qquad
\bar{\mathbf{B}} = \bigl(e^{\delta \mathbf{A}} - \mathbf{I}\bigr)\,\mathbf{A}^{-1}\mathbf{B}.
\]
By transforming the parameters from ($\delta$, $\mathbf{A}$, $\mathbf{B}$) to ($\bar{\mathbf{A}}$, $\bar{\mathbf{B}}$), the SSM model becomes a sequence-to-sequence mapping framework from discrete input to output.

Real-world dynamics are seldom linear time-invariant (LTI); their behaviour shifts with context, input, and time. As shown in Mamba \citep{gu2023mamba}, making the state-space model content‑varying allows the network to focus on relevant signals and better capture nonstationary processes. Mamba achieves this by modulating the SSM parameters with selective, data-dependent gates, yielding a context-aware adaptive transition. This is achieved by modifying the parameters as functions of the input sequence. 
For each token $u_i \in \mathbb{R}^{d}$ at step $i$ with $d$-dimensional, we compute adaptive parameters:
\begin{equation}
[\hat{\delta_i}, B_i, C_i] = W_{proj} u_i + \mathbf{b}_{proj}, \quad \delta_i=\operatorname{softplus}(\hat{\delta_i})
\label{eq:proj}
\end{equation}

Here, a single linear projection with weights $W_{proj}$and bias $b_{proj}$ of the input token $u_i\!\in\!\mathbb{R}^d$ produces three vectors: (1) $\hat{\delta_i}$: a step–size logit. After softplus, $\delta_i>0$ is used as the content‑dependent step in the discretized SSM update; (2) $B_i$: an input gate that scales the driving term of the state update; 
(3) $C_i$: an output/skip gate that scales the direct contribution of the input to the output.

\subsection{Formulation of Time-Aware $\Delta t$-Mamba3D}
Time‑Aware $\Delta t$‑Mamba3D aims to capture the
clinically important inter‑visit time gap and the spatial–temporal
dependencies among neighbouring latent states.  
Unlike earlier visual Mamba variants that rely on multiple scan
directions and content‑only adaptive steps, we introduce two key
modifications:  (i) the true time gap~$\Delta t$ is injected directly into the selective
scan, and (ii) a 3D neighborhood fusion term is added to the original Mamba
equations. The detailed workflow is illustrated in Fig. 2

\paragraph{Time-aware Modulation}
Mamba \cite{gu2023mamba} is designed for language modeling, implicitly assuming uniform steps between tokens (i.e., $\Delta t = 1$). To mimic variable dependence, it uses the current input $u_i$ as a control signal and predicts an effective step via
$\delta_i = \mathrm{softplus}(\mathrm{Linear}(u_i))$, selectively copying past inputs.  
However, this content-gated surrogate does not encode actual, irregular clock times—ubiquitous in clinical longitudinal data. We therefore augment the model with explicit time information, using real $\Delta t$ (exam gaps) alongside $u_i$ as control signals.
 Let $\Delta{t_i}$ denote the real calendar interval (in months) between the current time $t_i$ and its preceding time $t_{i-1}$, with a minimum interval $\tau_{\min}=12$ months for normalizing time gap. The time-aware step size $\Delta_i$ is defined by:
\begin{equation}
\delta_i^{\text{TA}}=\delta_i \left(1+\gamma \frac{\Delta{t(i)}}{\tau_{\min}}\right), \quad 0<\gamma \le 1.
\label{eq:dt-step}
\end{equation}
Thus, tokens within the same visit retain the original microscopic step $\delta_i$, whereas tokens at visit boundaries scale proportionally to the actual elapsed interval, which decide how much past information is carried forward. This ensures numerical stability and meaningful time-awareness, where $\gamma$ is a mixing coefficient that controls how strongly real time stretches the step size.
To theoretically demonstrate that the proposed time-aware encoding can control the importance between historical memory and current input, we give the following Theorem proven by \cite{li2024dyg}.
\paragraph{Theorem 1}
\textit{Let $A= V\Lambda V^{-1}$ with eigenvalues $\Lambda=\operatorname{diag}(\lambda_1,\dots,\lambda_N)$ and $\operatorname{Re}(\lambda_i)\le 0$. \begin{align}
\bar{A}_k &= \operatorname{diag}\!\left(e^{\lambda_1 \delta_i^{\text{TA}}},\,  \ldots,\, e^{\lambda_N \delta_i^{\text{TA}}}\right),\\[4pt]
\bar{B}_k &= \bigl(\lambda_1^{-1}\!\left(e^{\lambda_1 \delta_i^{\text{TA}}}-1\right),\,  \ldots,\, \lambda_N^{-1}\!\left(e^{\lambda_N \delta_i^{\text{TA}}}-1\right)\bigr).
\end{align}
The k-th coordinate-wise hidden of $h_i$ state update is:}
\begin{equation}
h_{k,i}=e^{\lambda_k\delta_i^{\text{TA}}}h_{k,i-1}+\lambda_k^{-1}(e^{\lambda_k\delta_i^{\text{TA}}}-1)u_{k,i}.
\label{eq:theorem}
\end{equation}

According to the Theorem, if $\delta_i^{\text{TA}}$
is small enough, we observe $h_{k,i}\approx h_{k,i-1}$, demonstrating that a short $\delta_i^{\text{TA}}$ arising from a small visit gap or uninformative content persists in the historical state that ignores the current input. And a larger $\delta_i^{\text{TA}}$ drives the $\lambda_k \delta_i^{\text{TA}} \ll 0$ and thus \( h_{k,i} \approx -\lambda_k^{-1} u_{k,i} \) meaning the contribution from the previous state is effectively forgotten and the update is dominated by the current input.

\paragraph{3D Neighborhood Fusion}
After selectively scanning all tokens by injecting time aware visit gap, if dependencies are still modeled along a single 1D order, it will leave residual spatial‑temporal interactions underexplored. We therefore apply a 3D neighborhood fusion step using depth-wise 3D convolutions aiming to capture the spatial and temporal dependencies of neighboring features in the latent state space:
\begin{equation}
\begin{aligned}
x_i &= \bar{A}\,x_{i-1} + \bar{B}\,u_i,\\
h_i &= \sum_{k\in\Omega} \alpha_k\, x_{\pi_k(i)}, \\y_i &= C_i\,h_i + D\,u_i,
\end{aligned}
\end{equation}
where $x_i$ is the original state variable, $h_i$ is the spatio-temporal aware  state variable, $\Omega$ is the neighbor set, $a_k$ is a learning weight, and $\pi_k$ indexes the 3D coordinates of the $k$-th neighbor. The original state variable $x_i$ is directly influenced by its previous state with newly added $\Delta t_i$ while the spatio-temporal aware variable $h_i$ incorporates additional neighboring state variables through a fusing mechanism. For each state $x_i$, we linearly weight its neighboring states $\pi_k(i)$ in $\Omega$ with coefficients $\alpha_k$ to integrate spatiotemporal context into a new state $h_i$. By considering both the global long-range and the local spatial and temporal information, the fused state variable gains a richer context.

To capture residual spatio‑temporal dependencies, we linearly re‑weight adjacent states with depth‑wise 3‑D convolutions.  
Given the maximum number of visits $T$, we employ two asymmetric kernels: $(1,3,3)$ to capture purely spatial context and $(\min(3,T),3,3)$ to capture joint spatio\textendash temporal information.

\[
\mathcal{K} = \Bigl\{ (1,3,3),\; (\min(3,T),\,3,\,3) \Bigr\}.
\]

and fuse their responses as
\begin{equation}
\mathbf{h}=\sum_{s\in\mathcal{K}} \beta_s\,\operatorname{DW‑Conv3D}^{(s)}\!\bigl(\hat{\mathbf{x}}\bigr),
\qquad
\beta_s=\frac{e^{\alpha_s}}{\displaystyle\sum_{s\in\mathcal{K}} e^{\alpha_{s}}}.
\label{eq:sa_fuse}
\end{equation}
Let $\hat{\mathbf{x}}\in\mathbb{R}^{d\times T\times H\times W}$ be the state tensor
reshaped to its spatio-temporal grid.  
We apply a depth-wise 3-D convolution
\(
\operatorname{DW\text{-}Conv3D}^{(s)}(\cdot)
\)
with kernel size $s$,  padding to the same size, and groups $=d$, so that each
of the $d$ channels is filtered independently.  
This preserves the spatial–temporal receptive field of a standard
Conv3D while reducing parameters and FLOPs by a factor of $d$. The $\alpha_s$ are learnable logits; $\beta_s$ are their softmax-normalized weights, ensuring $\beta_s\ge0$ and $\sum_s \beta_s=1$.

\begin{figure*}[t]
  \centering
  \includegraphics[width=0.8\linewidth]{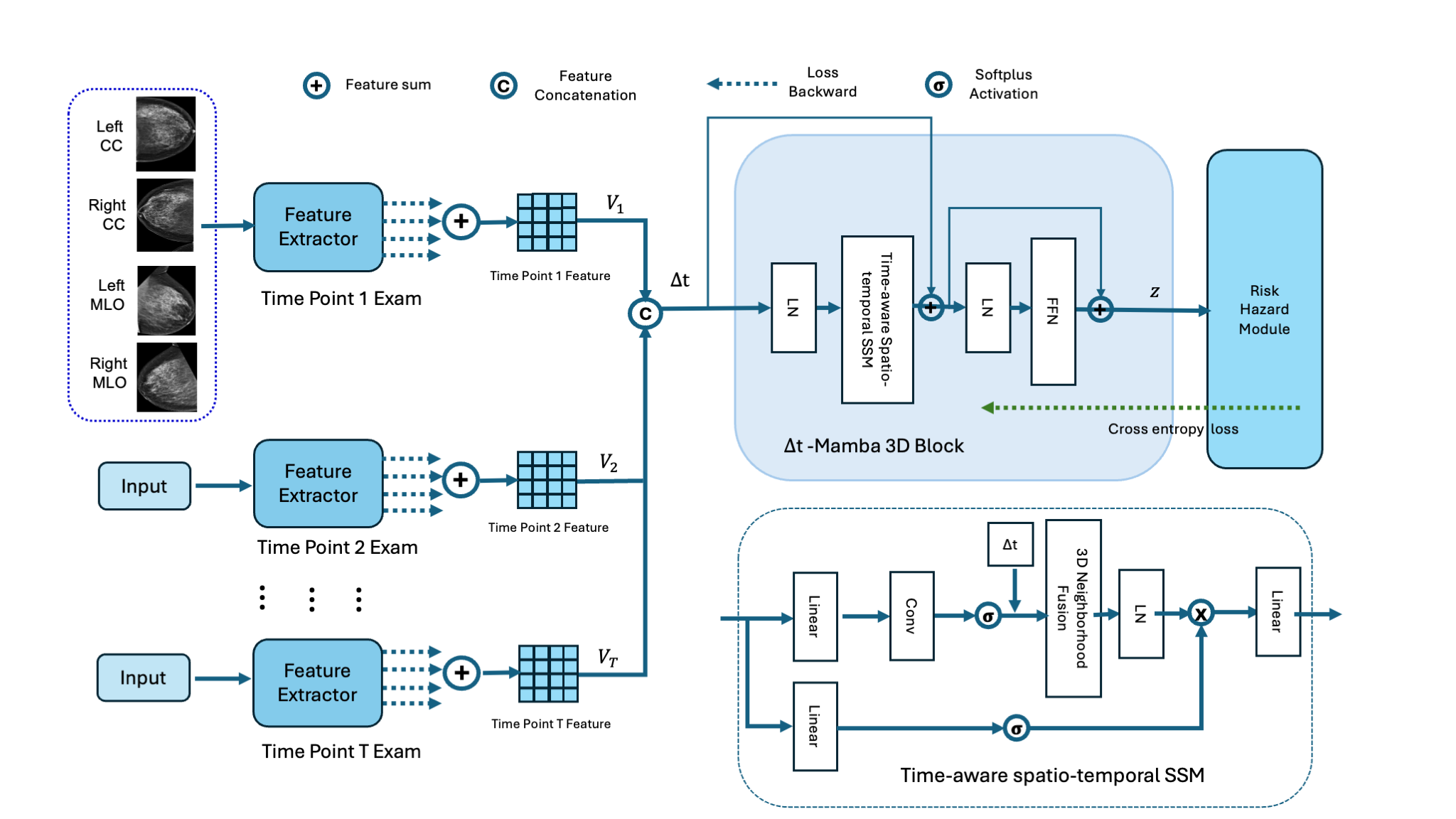}
  
  \caption{Overall architecture of the proposed Time-Aware $\Delta t$-Mamba3D.
  Top: longitudinal multi-visit, 4-view mammography is encoded per view with Swin-V2,
  fused by visit, and processed by a hierarchy of $\Delta t$-Mamba3D blocks before the Risk Hazard Module. Bottom right: Expanded diagram of a single $\Delta t$-Mamba3D block.}
  \label{fig:sa_dt_mamba3d_block}
\end{figure*}

\subsection{Model Architecture}
\label{sec:model-arch}

As shown in Fig. 3,  Each patient contributes a longitudinal series of imaging visits at
irregular times $0 < t_1 < \dots < t_T$.  At visit
$t$, a standard mammography study provides four projections/views:
left/right craniocaudal (L/R-CC) and left/right mediolateral oblique
(L/R-MLO).  Let $\mathbf{I}_{t,v}\!\in\!\mathbb{R}^{3\times H\times W}$
denote the $v$th view ($v\!\in\!\{1,\dots,4\}$). We process each image with a Swin-V2 backbone \citep{liu2021swin} that produces a low-resolution
feature map
\(
\mathbf{F}_{t,v} = \operatorname{Swin}(\mathbf{I}_{t,v})
 \in \mathbb{R}^{d \times H_0 \times W_0}
\)
($H_0,W_0\!\ll\!H,W$; e.g., $H_0{=}W_0{=}8$, $d{=}768$).  
Because radiologists integrate information across symmetric left/right
and CC/MLO views, we fuse per-visit features by summation, yielding a fused spatial tensor
\(
\mathbf{V}_t \in \mathbb{R}^{d \times H_0 \times W_0}.
\)
Stacking across valid visits gives
\(
\mathbf{V} \in \mathbb{R}^{d \times T \times H_0 \times W_0}.
\)
\paragraph{Time-Aware $\Delta t$-Mamba3D block.}
We apply the proposed spatio-temporal state-space block
 to $\mathbf{V}$ together with the inter-visit
gaps $\Delta t = t_i - t_{i-1}$ ($\Delta t_1=0$).  The block: (i) flattens the $T_0{\times}H_0{\times}W_0$ extracted feature to a token sequence,
(ii) runs a Mamba selective scan whose state update is modulated by the
true $\Delta t$ for all tokens belonging to visit $t$, and
(iii) reshapes back to 3D format and applies a learnable mixture of
different depth-wise $3$D kernels for
3D neighborhood fusion. 
The output of the block is
\(
\mathbf{Z} \in \mathbb{R}^{d \times T \times H \times W}.
\)

\paragraph{Patient embedding and Risk Hazard Module.}
After encoding the irregular time gaps and the spatio-temporal context, we aggregate $\mathbf{V}$ across space (mean pool over $H_0,W_0$) and time
(masked mean over the $T$ valid visits) to obtain a patient embedding
\(
\mathbf{z} \in \mathbb{R}^d.
\)
This embedding feeds the Risk Hazard Module.  In our breast cancer risk prediction setting we utilize an additive hazard \cite{yala2021toward, karaman2024longitudinal} and integrate the entire history embedding $z$, to estimate the future risk of developing breast cancer. The cumulative risk over \( k \) years (where \( k \in \{1, 2, \ldots, 5\}\), as we target to predict 1- to 5-year risk) is computed by summing the baseline risk $B_r(z)$ with the annual hazard term $H_i(z)$ \begin{equation}
P\!\left(t_{\text{cancer}} = k \mid z\right)
= \sigma\!\left( B_r(z) + \sum_{i=1}^{k} H_i(z) \right).
\label{eq:cancer_prob}
\end{equation}
\paragraph{Handling variable-length series.}
Because patients have different numbers of exams, we left-pad every sequence to a fixed maximum length~$T$. A binary
mask suppresses padded tokens in the temporal pooling step and supplies
the correct $\Delta t$ values (zero at padded positions).

\begin{table}[t]
\centering
\scriptsize          
\caption{Single-layer efficiency on a 512-token input (\(T=8,\;H=W=8\)) with hidden
width \(d=768\).  FLOPs are fused multiply–adds (billions).
``Peak tokens/s'' is an upper-bound estimate for an FP16 A100 (312 TFLOP/s) at
30\,\% utilization.}
\label{tab:layer-cost}
\begin{tabular}{lccc}
\toprule
\textbf{Model}                    & \textbf{Params (M)} & \textbf{FLOPs (G)} & \textbf{Peak tokens/s} (\(\times10^{6}\)) \\
\midrule
I3D (3D CNN) & 15.9 & 8.2  & 5.9  \\
Transformer          & 7.1 & 4.0  & 11.9 \\
GRU-\(\Delta t\)     & 3.5 & 0.06 & 79.0 \\
Neural ODE (6 steps)              & 1.2 & 0.36 & 56.5 \\
ContiFormer       & 7.5 & 4.1  & 11.6 \\
TimeSformer           & 7.1 & 2.4  & 19.6 \\
SegMamba           & 5.3 & 0.90 & 49.7 \\
LongMamba              & 5.4 & 0.90 & 49.7 \\
\textbf{\(\Delta t\)-Mamba3D (ours)} & \textbf{1.8} & \textbf{0.32} & \textbf{59.3} \\
\bottomrule
\end{tabular}

\vspace{2pt}

\end{table}
%



\begin{table*}[t]
\centering
\caption{Performance comparisons (mean $\pm$ std) on the \textbf{CSAW} (max 4 exams) and \textbf{Independent} (max 8 exams) datasets. 
All models share the same Swin-V2 per-visit encoder; model names indicate whether inter-visit gaps ($\Delta t$) are used.}
\scriptsize
\setlength{\tabcolsep}{3pt}
\begin{tabular}{l ccccc cccccc}
\toprule
& \multicolumn{5}{c}{\textbf{CSAW Dataset(max 4 prior exams)}} & \multicolumn{6}{c}{\textbf{Independent Dataset (max 8 prior exams)}}\\
\cmidrule(lr){2-6}\cmidrule(lr){7-12}
\textbf{Model} &
\textbf{c-idx} & \textbf{AUC$_{2y}$} & \textbf{AUC$_{3y}$} & \textbf{AUC$_{4y}$} & \textbf{AUC$_{5y}$} &
\textbf{c-idx} & \textbf{AUC$_{1y}$} & \textbf{AUC$_{2y}$} & \textbf{AUC$_{3y}$} & \textbf{AUC$_{4y}$} & \textbf{AUC$_{5y}$} \\
\midrule
\multicolumn{12}{l}{\textit{Time-aware (pooled) baselines}}\\
GRU-$\Delta t$                 & 0.661$\pm$0.03 & 0.677$\pm$0.01 & 0.661$\pm$0.01 & 0.647$\pm$0.01 & 0.643$\pm$0.01 & 0.576$\pm$0.01 & 0.620$\pm$0.01 & 0.610$\pm$0.01 & 0.587$\pm$0.01 & 0.573$\pm$0.01 & 0.571$\pm$0.02 \\
Transformer                    & 0.635$\pm$0.01 & 0.634$\pm$0.02 & 0.631$\pm$0.01 & 0.625$\pm$0.01 & 0.623$\pm$0.01 & 0.647$\pm$0.04 & 0.659$\pm$0.01 & 0.629$\pm$0.02 & 0.605$\pm$0.01 & 0.584$\pm$0.02 & 0.573$\pm$0.02 \\
Neural ODE                     & 0.649$\pm$0.01 & 0.657$\pm$0.01 & 0.656$\pm$0.01 & 0.658$\pm$0.01 & 0.658$\pm$0.01 & 0.642$\pm$0.02 & 0.651$\pm$0.03 & 0.652$\pm$0.04 & 0.641$\pm$0.02 & 0.626$\pm$0.01 & 0.609$\pm$0.03 \\
ContiFormer                    & 0.659$\pm$0.01 & 0.661$\pm$0.01 & 0.673$\pm$0.01 & 0.681$\pm$0.01 & 0.680$\pm$0.01 & 0.714$\pm$0.02 & 0.695$\pm$0.03 & 0.709$\pm$0.03 & 0.694$\pm$0.04 & 0.665$\pm$0.05 & 0.643$\pm$0.05 \\
$\Delta t$-Mamba   & 0.716$\pm$0.01 & 0.738$\pm$0.02 & 0.714$\pm$0.01 & 0.700$\pm$0.01 & 0.695$\pm$0.02 & 0.717$\pm$0.01 & 0.731$\pm$0.01 & 0.726$\pm$0.01 & 0.708$\pm$0.02 & 0.677$\pm$0.03 & 0.657$\pm$0.04 \\
\midrule
\multicolumn{12}{l}{\textit{Spatio-temporal baselines (uniform time)}}\\
I3D       & 0.690$\pm$0.01 & 0.700$\pm$0.01 & 0.688$\pm$0.02 & 0.680$\pm$0.02 & 0.673$\pm$0.02 & 0.712$\pm$0.01 & 0.722$\pm$0.01 & 0.717$\pm$0.02 & 0.683$\pm$0.02 & 0.654$\pm$0.02 & 0.636$\pm$0.01 \\
TimeSformer       & 0.662$\pm$0.01 & 0.679$\pm$0.01 & 0.674$\pm$0.01 & 0.636$\pm$0.02 & 0.633$\pm$0.02 & 0.717$\pm$0.01 & 0.725$\pm$0.01 & 0.709$\pm$0.01 & 0.687$\pm$0.02 & 0.665$\pm$0.02 & 0.643$\pm$0.04 \\
SegMamba              & 0.704$\pm$0.01 & 0.732$\pm$0.03 & 0.692$\pm$0.01 & 0.684$\pm$0.01 & 0.672$\pm$0.01 & 0.714$\pm$0.02 & 0.723$\pm$0.02 & 0.719$\pm$0.01 & 0.709$\pm$0.01 & 0.688$\pm$0.01 & 0.658$\pm$0.01 \\
LongMamba & 0.711$\pm$0.02 & 0.722$\pm$0.03 & 0.707$\pm$0.02 & 0.698$\pm$0.02 & 0.689$\pm$0.03 & 0.712$\pm$0.01 & 0.695$\pm$0.01 & 0.695$\pm$0.02 & 0.671$\pm$0.02 & 0.656$\pm$0.01 & 0.621$\pm$0.01 \\
Mamba3D      & 0.713$\pm$0.02 & 0.737$\pm$0.02 & 0.707$\pm$0.02 & 0.710$\pm$0.02 & 0.708$\pm$0.03 & 0.716$\pm$0.03 & 0.729$\pm$0.03 & 0.719$\pm$0.02 & 0.715$\pm$0.01 & 0.698$\pm$0.04 & 0.679$\pm$0.02 \\
\midrule
\textbf{$\Delta t$-Mamba3D} &
\textbf{0.742$\pm$0.01} & \textbf{0.754$\pm$0.02} & \textbf{0.743$\pm$0.02} & \textbf{0.730$\pm$0.01} & \textbf{0.720$\pm$0.01} &
\textbf{0.738$\pm$0.01} & \textbf{0.749$\pm$0.02} & \textbf{0.752$\pm$0.02} & \textbf{0.733$\pm$0.03} & \textbf{0.719$\pm$0.04} & \textbf{0.705$\pm$0.05} \\
\bottomrule
\end{tabular}
\label{tab:main_compare_max4_8}
\end{table*}

\begin{table*}[t]
\centering
\caption{Ablation study on the spatial--temporal module design on the CSAW and Independent datasets. Values are mean $\pm$ std.}
\scriptsize
\setlength{\tabcolsep}{2pt}
\begin{tabular}{l ccccc cccccc}
\toprule
& \multicolumn{5}{c}{\textbf{CSAW Dataset(max 4 prior exams)}} & \multicolumn{6}{c}{\textbf{Independent Dataset (max 8 prior exams)}}\\
\cmidrule(lr){2-6}\cmidrule(lr){7-12}
\textbf{Model} &
\textbf{c-idx} & \textbf{AUC$_{2y}$} & \textbf{AUC$_{3y}$} & \textbf{AUC$_{4y}$} & \textbf{AUC$_{5y}$} &
\textbf{c-idx} & \textbf{AUC$_{1y}$} & \textbf{AUC$_{2y}$} & \textbf{AUC$_{3y}$} & \textbf{AUC$_{4y}$} & \textbf{AUC$_{5y}$} \\
\midrule
Baseline                     & 0.701$\pm$0.02 & 0.726$\pm$0.02 & 0.694$\pm$0.05 & 0.680$\pm$0.05 & 0.670$\pm$0.06 & 0.688$\pm$0.01 & 0.699$\pm$0.03 & 0.697$\pm$0.01 & 0.676$\pm$0.01 & 0.660$\pm$0.01 & 0.642$\pm$0.02  \\
$k=\{1,3,3\}$                & 0.705$\pm$0.01 & 0.730$\pm$0.01 & 0.709$\pm$0.01 & 0.699$\pm$0.01 & 0.697$\pm$0.01 & 0.710$\pm$0.03 & 0.724 $\pm$0.04 & 0.702$\pm$0.02 & 0.694$\pm$0.02 & 0.688$\pm$0.01 & 0.649$\pm$0.01 \\
$k=\{3,3,3\}$                & 0.702$\pm$0.01 & 0.725$\pm$0.01 & 0.695$\pm$0.02 & 0.692$\pm$0.01 & 0.682$\pm$0.00 & 0.718$\pm$0.01 & 0.718$\pm$0.01 & 0.715$\pm$0.01 & 0.702$\pm$0.01 & 0.681$\pm$0.02 & 0.661$\pm$0.02 \\
$k=\{1\&3,3,3\}$              & 0.713$\pm$0.02 & 0.737$\pm$0.02 & 0.707$\pm$0.02 & 0.710$\pm$0.02 & 0.708$\pm$0.03 & 0.716$\pm$0.03 & 0.729$\pm$0.03 & 0.719$\pm$0.02 & 0.715$\pm$0.01 & 0.698$\pm$0.04 & 0.679$\pm$0.02 \\
\midrule
Inter-Slice $\Delta t$-Mamba     & 0.682$\pm$0.01 & 0.691$\pm$0.01 & 0.658$\pm$0.01 & 0.657$\pm$0.01 & 0.662$\pm$0.02 & 0.692$\pm$0.01 & 0.707$\pm$0.01 & 0.706$\pm$0.02 & 0.697$\pm$0.05 & 0.675$\pm$0.06 &
0.656$\pm$0.06 \\
$\Delta t$-Mamba             & 0.716$\pm$0.01 & 0.738$\pm$0.02 & 0.714$\pm$0.01 & 0.700$\pm$0.01 & 0.695$\pm$0.02 & 0.717$\pm$0.01 & 0.731$\pm$0.01 & 0.726$\pm$0.01 & 0.708$\pm$0.02 & 0.677$\pm$0.03 & 0.657$\pm$0.04 \\

\textbf{$\Delta t$-Mamba3D} &
\textbf{0.742$\pm$0.01} & \textbf{0.754$\pm$0.02} & \textbf{0.743$\pm$0.02} & \textbf{0.730$\pm$0.01} & \textbf{0.720$\pm$0.01} &
\textbf{0.738$\pm$0.01} & \textbf{0.749$\pm$0.02} & \textbf{0.752$\pm$0.02} & \textbf{0.733$\pm$0.03} & \textbf{0.719$\pm$0.04} & \textbf{0.705$\pm$0.05} \\
\bottomrule
\end{tabular}
\label{tab:main_compare_max4_8}
\end{table*}

\section{Experiments}
\subsection{Study Cohorts and Datasets}
Our experiments used two independent patient cohorts and imaging datasets. The first is the Karolinska Case-Control \textbf{(CSAW-CC) Dataset} \cite{Strand2022}, which is a part of the Cohort of Screen-Aged Women (CSAW). The CSAW-CC dataset was specifically curated for developing breast imaging-based AI tools. It includes women aged 40–74 years old who underwent mammographic screening between 2008 and 2016 using Hologic imaging systems. To mitigate potential bias in the risk prediction due to early cancer signs or early-detectable cancers, patients diagnosed with breast cancer within six months following the “present” screening exam were excluded. Our analysis included subjects who have at least two sequential screening exams. The final CSAW-CC cohort consisted of 406 breast cancer cases (all biopsy-proven) and 6,053 normal controls, with inter-exam intervals ranging from 12 to 36 months. The second dataset (denoted as \textbf{Independent Dataset}) is a retrospectively collected case-control cohort at a different hospital, with individuals who participated in routine breast cancer screening from 2007 to 2014 also using Hologic systems. We have data use agreement for this not-publicly-available dataset. This cohort comprises 293 breast cancer cases (all biopsy-proven) and 297 normal controls (at least 1-year follow-up to ensure normal status). Each subject had at least two sequential screening mammogram exams, with inter-exam intervals ranging from 12 to 24 months. Detailed dataset descriptions are provided in Appendix.

\subsection{Implementation Details}

$\Delta t$-Mamba3d models were trained to predict 1- to 5-year breast cancer risk using sequential screening mammograms. For each mammogram exam of a patient’s data, it is treated as a reference time-point (Prior 0) and we then traced backward up to maximum three prior exams in CSAW dataset and up to seven in the Independent Dataset, with irregular intervals of 12–36 months between consecutive exams. Patient outcomes (e.g., cancer vs. normal status) were determined based on the next follow-up exam occurring after $k$ years since Prior 0, where $k$ corresponds to the prediction horizon (1–5 years) \cite{yala2021toward}. All dataset splits were rigorously performed at the patient level to prevent data leakage.

We employed patient-wise 5-fold cross-validation to evaluate the performance of the proposed $\Delta t$-Mamba3D model on both datasets. In each fold, data is split into  training and testing set in an 80\%-20\% ratio. To focus the model learning on breast tissue, we first used LIBRA \cite{keller2012estimation,keller2015preliminary} to segment the breasts and discard the background, producing images of size $350\times400$ pixels. To mitigate class imbalance in the CSAW dataset, we adopt the reweighted cross-entropy loss function. The model was trained for 30 epochs with a batch size of 8, and the best checkpoint was selected via a grid search over learning rate of 5e‑5 and 1e‑5. All experiments were conducted on an NVIDIA TESLA A100 GPU, courtesy of our institution's computing resources. Model performance was evaluated using C-index and the Area Under the ROC Curve (AUC), with the mean AUC and standard deviations computed over 5-fold cross-validation for predicting the 1- to 5-year risk. 

We compared our method with state-of-the-art methods including: (i) vector-based time-series methods—GRU-$\Delta t$\cite{che2018recurrent},  vanilla Transformer\cite{vaswani2017attention}, Neural ODE\cite{rubanova2019latent}, and ContiFormer\cite{chen2023contiformer}; For benchmarking, each exam is associated with a single inter‑visit gap $\Delta t$, which we feed to GRU‑$\Delta t$, Neural ODE, and ContiFormer while leaving all other model components unaltered. We grid‑searched hidden sizes $\{256,\,512\}$ and depths $\{1\text{--}3\}$ to give each model its best setting. (ii) spatio-temporal approaches-TimeSformer\cite{bertasius2021space}, SegMamba \cite{xing2024segmamba}, and LongMamba\cite{zhou2025longmambattn}. Two variants of our proposed framework were also included for comparison: Mamba3D, which uses only 3D neighborhood fusion, and $\Delta$t-Mamba, a time-aware state-space model without 3D neighborhood fusion. In addition, we also compared performance of our model to several representative longitudinal breast cancer risk models: Mirai \cite{yala2021toward}, LRP-NET \cite{dadsetan2022deep}, Prime+ \cite{lee2023enhancing}, and LoMaR \cite{karaman2024longitudinal}. Detailed comparison results with these related models are provided in the Appendix. Finally, we compared the computational complexity of different methods.

\section{Results}
\label{sec:results}

Table 2 reports validation performance of $\Delta t$-Mamba3D along with compared methods under two settings: using up to four prior exams on CSAW Dataset and up to eight exams on Independent Dataset for breast cancer risk prediction . All models utilized the same Swin-V2 per-visit encoder, differing primarily in their handling of inter-visit intervals ($\Delta t$) and in capturing spatio-temporal information. 

For the CSAW dataset, time-aware pooled baselines such as GRU-$\Delta t$, Transformer, Neural ODE, ContiFormer, and $\Delta t$-Mamba achieved C-indices ranging from 0.635 to 0.716. Among these, $\Delta t$-Mamba exhibited the highest performance with a C-index of 0.716 and notable AUC scores across all evaluated intervals. Spatio-temporal baselines, including I3D, TimeSformer, SegMamba, LongMamba, and Mamba3D, showed improved performance, with Mamba3D achieving a C-index of 0.713. Our proposed $\Delta t$-Mamba3D significantly outperformed all baselines, achieving the highest C-index of 0.742 and superior AUC performance consistently across 2-year (0.754), 3-year (0.743), 4-year (0.730), and 5-year (0.720) risk prediction.

For the Independent Dataset, similar trends were observed. The time-aware pooled baselines achieved moderate performance, with $\Delta t$-Mamba reaching the highest C-index of 0.714. Spatio-temporal baselines generally showed improvements, with Mamba3D  obtaining a notable C-index of 0.716. Our proposed $\Delta t$-Mamba3D model again delivered the highest performance, yielding a C-index of 0.738 and showing consistently superior AUC scores across all intervals evaluated, including 1-year (0.749), 2-year (0.752), 3-year (0.733), 4-year (0.719), and 5-year (0.705). Overall, these results underscore the effectiveness of our proposed method by incorporating true inter-visit intervals ($\Delta t$) and spatio-temporal information into the $\Delta t$-Mamba3D architecture, highlighting significant improvements over existing state-of-the-art methods, particularly on long-sequence data.

\subsection{Computational Complexity}
\label{sec:complexity}
As shown in Table 1, \textbf{$\Delta t$‑Mamba3D} delivers an unrivalled efficiency trade‑off.  
With only 1.8\,M parameters and 0.32\,GFLOPs per 512‑token layer, it achieves a
throughput of 59.3\,M tokens/s— much lighter and 
faster than a vanilla Transformer and I3D.  It also surpasses
other attention-based TimeSformer and ContiFormer by
3--5$\times$ in speed while using far less compute.  Compared with its own family variants, SegMamba and LongMamba, it trims about two‑thirds of their parameters and FLOPs yet is still at least 20\% faster.  Although the recurrent GRU‑$\Delta t$ baseline records the highest raw speed (79.0\,M tokens/s), it requires twice as many parameters and lacks 3‑D spatial modelling depth, while the smaller Neural ODE matches throughput only at a slightly higher FLOP budget.  

\subsection{Ablation Study}
In this section, we ablate several key components of our proposed method to evaluate effects by using the CSAW and Independent datasets. We construct the baseline model using the original mamba. 1) \textbf{Neighbor set}: We introduce two depth‑wise 3‑D branches: a purely spatial filter \((1\times3\times3)\) and a spatio‑temporal filter \((3\times3\times3)\).
Either branch alone improves upon the baseline, and using them together yields an even larger overall performance gain.
  \textbf{Time-aware module}: Inter‑Slice $\Delta t$-Mamba denotes alternating patches across visits (inter‑slice scan) while injecting the true time gap at every hop. This variant performs poorly because it (i) breaks spatial coherence, and (ii) repeatedly triggers $\Delta t$ gates, adding noise and destabilizing learning. Adding $\Delta t$ alone raises the model performance, and the full $\Delta t$-Mamba3D achieves the top scores across all horizons.
Overall observations: (i) coupling the spatial-only branch with the spatio‑temporal branch delivers the best results, because each captures complementary cues. 
In our setting, scanning an image intra-slice before tokenization preserves these advantages and further boosts performance; (ii) coupling time-aware scanning with 3D neighborhood  fusion yields the most effective spatio-temporal integration.

\section{Conclusion}
\label{sec:conclusion}

We introduced Time-Aware $\boldsymbol{\Delta t}$\mbox{-}Mamba3D, a 
spatial--temporal state--space block for modeling longitudinal breast imaging.
This method modulates each state update by the true inter-visit interval $\Delta t$, and applies a learnable multiscale, depth‑wise 3D neighborhood fusion module that jointly models spatio-temporal structure and irregular temporal patterns.  Integrated into a
breast cancer risk prediction framework (up to eight sequential imaging exams/visits; four
views per exam), our model consistently outperforms recurrent,
transformer, and prior visual SSM baselines in $c$-index and
long-horizon AUCs (1--5y) under two different datasets. Our proposed method scales
linearly in sequence length, enabling computationally efficient modeling of decade-long patient
histories where quadratic attention methods like transformer become impractical. Future work will focus on extending the proposed method to additional medical imaging modalities and related clinical tasks.

\section{Acknowledgments}
This work was supported in part by a National Institutes of Health (NIH) Other Transaction research contract~\#1OT2OD037972-01; a National Science Foundation (NSF) grant (CICI: SIVD; \#2115082); grant~1R01EB032896 as part of the NSF/NIH Smart Health and Biomedical Research in the Era of Artificial Intelligence and Advanced Data Science Program; an Amazon Machine Learning Research Award; and the University of Pittsburgh Momentum Funds (Scaling Grant) for the Pittsburgh Center for AI Innovation in Medical Imaging.
This work used Bridges-2 at the Pittsburgh Supercomputing Center through allocation [MED200006] from the Advanced Cyberinfrastructure Coordination Ecosystem: Services \& Support (ACCESS) program, which is supported by NSF grants \#2138259, \#2138286, \#2138307, \#2137603, and \#2138296.
This research was also supported in computing resources by the University of Pittsburgh Center for Research Computing and Data (RRID: SCR\_022735) through the resources provided by the H2P cluster, which is supported by NSF award OAC-2117681.
The views and conclusions contained in this document are those of the authors and should not be interpreted as representing official policies, either expressed or implied, of the NIH or NSF.

\bibliography{aaai2026}
\clearpage


\section*{Supplementary material}

# Appendix

## S1 Dataset Statistics

**Table S1** The two mammogram imaging datasets (CSAW and an Independent cohort) exhibit distinct lead-time distributions and characteristics. The CSAW cohort lacks cancer exams within the 0–1 year interval but includes increasing numbers of screening exams as the interval extends (e.g., 51 exams for 1–2 years, peaking at 292 for 2–3 years). Conversely, the Independent cohort has more evenly distributed exams, starting from 284 within 0–1 year.

Patient characteristics also differ significantly between datasets: the CSAW dataset has shorter follow-up durations ($4.12 \pm 1.60$ years for cancer cases; $5.0 \pm 1.6$ years for controls) and fewer average visits ($2.79 \pm 0.81$ for cancer cases; $3.48 \pm 0.86$ for controls), while the Independent cohort has longer follow-up ($4.40 \pm 1.94$ years for cancer cases; $4.23 \pm 2.16$ years for controls) and a higher number of visits ($6.4 \pm 1.80$ for cancer cases; $5.1 \pm 2.1$ for controls).

**Rationale for the Independent Dataset.** This dataset is not publicly available; our access is granted for research purposes under a data use agreement. Detailed information (e.g., institution name) will be provided in the camera-ready version upon acceptance of the manuscript, consistent with the agreement's terms. We selected this dataset because of its unique strength of having up to eight longitudinal mammogram exams per patient. In general, longitudinal breast imaging datasets are scarce in the public domain. To the best of our knowledge, no publicly available mammogram imaging datasets offer an average of more than five visits/exams per patient. As our objective is to evaluate effects of a longer history of imaging exams for predicting breast cancer risk using our proposed method, this Independent Dataset with up to eight exams provides us a strong setting to conduct experiments and evaluate our method, along with the CSAW dataset. Moreover, the two datasets have different label distributions, so our choice of using these two datasets better represent real-world conditions to assess our method's robustness.

Table S1: Patient cohort characteristics and lead-time distribution for the CSAW and Independent Datasets. Numeric entries are mean $\pm$ SD or sample size, $n$ (%).

| Characteristic | CSAW | | Independent | |
|---|---|---|---|---|
| | Cancer (n = 406) | Controls (n = 6 053) | Cancer (n = 289) | Controls (n = 291) |
| Follow-up span (years) | $4.12 \pm 1.60$ | $5.0 \pm 1.6$ | $4.40 \pm 1.94$ | $4.23 \pm 2.16$ |
| Number of visits (exams) | $2.79 \pm 0.81$ | $3.48 \pm 0.86$ | $6.4 \pm 1.80$ | $5.1 \pm 2.1$ |
| **Age category** | | | | |
| < 56 years | 170 (41.9) | 3364 (55.6) | 126 (43.6) | 163 (56.4) |
| >= 56 years | 236 (58.1) | 2689 (44.4) | 163 (56.4) | 128 (43.6) |
| **Lead-time distribution (cancer cases only)** | | | | |
| Years to cancer | CSAW | | Independent | |
| 0–1 year | 0 | | 284 | |
| 1–2 years | 51 | | 268 | |
| 2–3 years | 292 | | 238 | |
| 3–4 years | 103 | | 190 | |
| 4–5 years[†] | 202 | | 168 | |

# S2 Comparison of the Proposed Risk Model with Existing Models

**Table S2** Table S2 summarizes the prediction performance comparisons of our model to various existing breast cancer risk prediction models on the two datasets. The proposed $\Delta t$-Mamba3D model consistently achieves the highest performance across both datasets and all metrics (C-index and 1- to 5-year AUC), outperforming other state-of-the-art longitudinal models including Mirai, LRP-NET, LoMaR, and Prime+. In order to make a consistent setting on all these models for comparisons, maximum 4 exams were used in the experiments. These experiment results are significant, not only showing that our method has technical strengths as seen in the main results of the paper, but also highlighting the higher clinical performance of our risk model than other existing models in predicting breast cancer risk.

Table S2: Prediction performance comparisons of the proposed model to other representative breast cancer risk prediction models on two datasets.

| Model | C-index | 1-year AUC | 2-year AUC | 3-year AUC | 4-year AUC | 5-year AUC |
|---|---|---|---|---|---|---|
| **CSAW-CC Dataset** | | | | | | |
| Mirai Yala et al. (2021) | $0.687 \pm 0.02$ | $--$ | $0.680 \pm 0.02$ | $0.664 \pm 0.03$ | $0.664 \pm 0.02$ | $0.630 \pm 0.01$ |
| LRP-NET Dadsetan et al. (2022) | $0.670 \pm 0.02$ | $--$ | $0.688 \pm 0.02$ | $0.654 \pm 0.01$ | $0.654 \pm 0.01$ | $0.636 \pm 0.02$ |
| LoMaR Karaman et al. (2024) | $0.696 \pm 0.01$ | $--$ | $0.708 \pm 0.01$ | $0.689 \pm 0.02$ | $0.686 \pm 0.03$ | $0.675 \pm 0.03$ |
| Prime+ Lee et al. (2023) | $0.683 \pm 0.02$ | $--$ | $0.697 \pm 0.01$ | $0.685 \pm 0.01$ | $0.681 \pm 0.01$ | $0.672 \pm 0.01$ |
| $\Delta t$-Mamba3D | $0.742 \pm 0.01$ | $--$ | $0.754 \pm 0.02$ | $0.743 \pm 0.02$ | $0.730 \pm 0.01$ | $0.720 \pm 0.01$ |
| **Independent Dataset** | | | | | | |
| Mirai Yala et al. (2021) | $0.685 \pm 0.02$ | $0.685 \pm 0.03$ | $0.660 \pm 0.03$ | $0.670 \pm 0.02$ | $0.655 \pm 0.03$ | $0.653 \pm 0.02$ |
| LRP-NET Dadsetan et al. (2022) | $0.676 \pm 0.02$ | $0.683 \pm 0.03$ | $0.665 \pm 0.03$ | $0.645 \pm 0.03$ | $0.640 \pm 0.02$ | $0.630 \pm 0.03$ |
| LoMaR Karaman et al. (2024) | $0.706 \pm 0.02$ | $0.717 \pm 0.02$ | $0.691 \pm 0.02$ | $0.661 \pm 0.03$ | $0.634 \pm 0.02$ | $0.620 \pm 0.02$ |
| Prime+ Lee et al. (2023) | $0.699 \pm 0.02$ | $0.702 \pm 0.01$ | $0.679 \pm 0.03$ | $0.644 \pm 0.03$ | $0.618 \pm 0.03$ | $0.597 \pm 0.02$ |
| $\Delta t$-Mamba3D | $0.746 \pm 0.02$ | $0.751 \pm 0.02$ | $0.733 \pm 0.02$ | $0.699 \pm 0.02$ | $0.676 \pm 0.03$ | $0.661 \pm 0.03$ |

# S3 Additional Experiment Results

**Table S3** To assess whether $\Delta t$-gating benefits other Mamba variants (e.g., LongMamba), we integrated it and evaluated on both datasets, using up to four exams per subject. Specifically, the CSAW dataset shows a notable improvement in c-index from 0.711 to 0.730, while the Independent dataset presents an even greater increase from 0.712 to 0.737. Similar improvements are observed across all evaluated time horizons (AUC metrics), affirming the efficacy of time-aware gating in enhancing longitudinal prediction accuracy.

**Table S4** Table S4 shows the validation performance of various models for predicting breast cancer risk using datasets limited to at most two prior exams. Among the evaluated models, the proposed $\Delta t$-Mamba3D consistently achieves the highest performance across both CSAW and Independent datasets, highlighting the advantage of explicitly incorporating temporal intervals and 3D spatial-temporal fusion in short-sequence scenarios. Compared with the main results in the paper that incorporates additional number of exams, our proposed model improves notably on long-horizon ($\geq$ 2-year) risk prediction, whereas the compared models show decreases in C-index and 1–5-year AUC. This indicates greater robustness of our method to a longer history of prior exams for risk prediction.

**Table S5** Parameter robustness—$\gamma$ ablation. Across CSAW (max 4 exams) and Independent (max 8 exams), we ablated the mixing ratio $\gamma$ that weights the content-aware and time-aware components. On CSAW, all mixtures ($\gamma \in \{0.2, 0.5, 1\}$) consistently outperformed both single-source baselines on c-index and AUCs across all horizons, with negligible sensitivity to $\gamma$. On the Independent dataset, the time-aware baseline already exceeded content-aware, and mixing further improved performance—$\gamma = 0.5$ produced the highest AUCs (1–5y) while $\gamma = 1$ slightly increased the c-index. For all experiments, we used $\gamma = 0.5$ except where noted.

Table S3: LongMamba with and without $\Delta t$ gating on two datasets (mean $\pm$ std).

| Dataset | Gating | c-index | $AUC_{1y}$ | $AUC_{2y}$ | $AUC_{3y}$ | $AUC_{4y}$ | $AUC_{5y}$ |
|---|---|---|---|---|---|---|---|
| CSAW | $\Delta t$ | 0.730 $\pm$ 0.01 | – | 0.752 $\pm$ 0.02 | 0.707 $\pm$ 0.01 | 0.705 $\pm$ 0.02 | 0.700 $\pm$ 0.02 |
|  | None | 0.711 $\pm$ 0.02 | – | 0.722 $\pm$ 0.03 | 0.707 $\pm$ 0.02 | 0.698 $\pm$ 0.02 | 0.689 $\pm$ 0.03 |
| Independent | $\Delta t$ | 0.737 $\pm$ 0.02 | 0.745 $\pm$ 0.02 | 0.729 $\pm$ 0.02 | 0.689 $\pm$ 0.03 | 0.667 $\pm$ 0.03 | 0.661 $\pm$ 0.03 |
|  | None | 0.712 $\pm$ 0.01 | 0.705 $\pm$ 0.01 | 0.695 $\pm$ 0.01 | 0.671 $\pm$ 0.02 | 0.656 $\pm$ 0.01 | 0.621 $\pm$ 0.01 |

Table S4: Validation performance (mean $\pm$ std ) for when using max 2 exams in the **CSAW** and **Independent** datasets. All models share the same Swin-V2 per-visit encoder;

| Model | CSAW (max 2 exams) | | | | | Independent (max 2 exams) | | | | | |
|---|---|---|---|---|---|---|---|---|---|---|---|
|  | c-idx | $AUC_{2y}$ | $AUC_{3y}$ | $AUC_{4y}$ | $AUC_{5y}$ | c-idx | $AUC_{1y}$ | $AUC_{2y}$ | $AUC_{3y}$ | $AUC_{4y}$ | $AUC_{5y}$ |
| *Time-aware (pooled) baselines* | | | | | | | | | | | |
| GRU-$\Delta t$ | 0.690$\pm$0.03 | 0.709$\pm$0.01 | 0.682$\pm$0.01 | 0.668$\pm$0.01 | 0.665$\pm$0.01 | 0.704$\pm$0.03 | 0.700$\pm$0.04 | 0.699$\pm$0.03 | 0.685$\pm$0.01 | 0.654$\pm$0.01 | 0.624$\pm$0.03 |
| Transformer | 0.640$\pm$0.02 | 0.635$\pm$0.02 | 0.641$\pm$0.02 | 0.638$\pm$0.01 | 0.638$\pm$0.01 | 0.702$\pm$0.01 | 0.709$\pm$0.01 | 0.701$\pm$0.02 | 0.688$\pm$0.00 | 0.664$\pm$0.01 | 0.640$\pm$0.02 |
| Neural ODE | 0.661$\pm$0.03 | 0.676$\pm$0.02 | 0.661$\pm$0.01 | 0.647$\pm$0.01 | 0.643$\pm$0.01 | 0.688$\pm$0.01 | 0.691$\pm$0.02 | 0.684$\pm$0.02 | 0.662$\pm$0.01 | 0.635$\pm$0.01 | 0.604$\pm$0.01 |
| ContiFormer | 0.717$\pm$0.01 | 0.722$\pm$0.01 | 0.715$\pm$0.01 | 0.699$\pm$0.02 | 0.695$\pm$0.02 | 0.726$\pm$0.00 | 0.725$\pm$0.01 | 0.722$\pm$0.00 | 0.685$\pm$0.00 | 0.651$\pm$0.01 | 0.626$\pm$0.01 |
| $\Delta t$-Mamba | 0.718$\pm$0.02 | 0.729$\pm$0.03 | 0.721$\pm$0.02 | 0.710$\pm$0.02 | 0.704$\pm$0.02 | 0.734$\pm$0.01 | 0.730$\pm$0.01 | 0.728$\pm$0.01 | 0.688$\pm$0.02 | 0.667$\pm$0.01 | 0.642$\pm$0.00 |
| *Spatio-temporal baselines (uniform time)* | | | | | | | | | | | |
| I3D | 0.705$\pm$0.01 | 0.722$\pm$0.02 | 0.709$\pm$0.01 | 0.692$\pm$0.02 | 0.692$\pm$0.01 | 0.712$\pm$0.03 | 0.728$\pm$0.04 | 0.719$\pm$0.03 | 0.705$\pm$0.01 | 0.664$\pm$0.01 | 0.634$\pm$0.02 |
| TimeSformer | 0.707$\pm$0.01 | 0.730$\pm$0.01 | 0.703$\pm$0.01 | 0.696$\pm$0.02 | 0.684$\pm$0.01 | 0.729$\pm$0.01 | 0.726$\pm$0.01 | 0.707$\pm$0.01 | 0.703$\pm$0.01 | 0.666$\pm$0.03 | 0.643$\pm$0.02 |
| SegMamba | 0.721$\pm$0.03 | 0.743$\pm$0.03 | 0.713$\pm$0.03 | 0.708$\pm$0.03 | 0.703$\pm$0.03 | 0.728$\pm$0.01 | 0.738$\pm$0.01 | 0.720$\pm$0.01 | 0.704$\pm$0.01 | 0.664$\pm$0.01 | 0.626$\pm$0.01 |
| LongMamba | 0.720$\pm$0.01 | 0.747$\pm$0.02 | 0.716$\pm$0.01 | 0.708$\pm$0.01 | 0.701$\pm$0.01 | 0.705$\pm$0.02 | 0.718$\pm$0.02 | 0.702$\pm$0.02 | 0.681$\pm$0.04 | 0.649$\pm$0.05 | 0.628$\pm$0.07 |
| Mamba3D | 0.722$\pm$0.02 | 0.746$\pm$0.02 | 0.720$\pm$0.02 | 0.709$\pm$0.02 | 0.703$\pm$0.02 | 0.733$\pm$0.01 | 0.740$\pm$0.01 | 0.721$\pm$0.01 | 0.701$\pm$0.01 | 0.666$\pm$0.03 | 0.644$\pm$0.04 |
| $\Delta t$-Mamba3D | 0.735$\pm$0.02 | 0.756$\pm$0.02 | 0.725$\pm$0.02 | 0.717$\pm$0.02 | 0.707$\pm$0.02 | 0.743$\pm$0.01 | 0.749$\pm$0.01 | 0.732$\pm$0.01 | 0.707$\pm$0.01 | 0.674$\pm$0.03 | 0.655$\pm$0.04 |

# S4 Time and Memory Complexity

We analyze a single $\Delta t$-Mamba3D layer for time and memory complexity. Let $L = T \times H_0 \times W_0$ be the sequence length, $d$ the feature dimension, $2d$ the expanded state dimension and set the SSM state dimension to $d_{\text{ssm}} = 16$.

Table S5: Robustness study on $\gamma$ for mixing time and content signals on the **CSAW** max 4 exams and **Independent** Datasets max 8 exams. All models share the same Swin-V2 per-visit encoder.

| | CSAW (max 4 exams) | | | | | Independent (max 8 exams) | | | | | |
|---|---|---|---|---|---|---|---|---|---|---|---|
| Model | c-idx | $AUC_{2y}$ | $AUC_{3y}$ | $AUC_{4y}$ | $AUC_{5y}$ | c-idx | $AUC_{1y}$ | $AUC_{2y}$ | $AUC_{3y}$ | $AUC_{4y}$ | $AUC_{5y}$ |
| Content-aware | 0.701±0.02 | 0.726±0.02 | 0.694±0.05 | 0.680±0.05 | 0.670±0.06 | 0.688±0.01 | 0.699±0.03 | 0.697±0.01 | 0.676±0.01 | 0.660±0.01 | 0.642±0.02 |
| Time-aware | 0.705±0.01 | 0.722±0.02 | 0.705±0.01 | 0.697±0.01 | 0.681±0.01 | 0.715±0.01 | 0.705±0.02 | 0.711±0.01 | 0.697±0.02 | 0.667±0.02 | 0.652±0.03 |
| $\gamma$=0.2 | 0.704±0.02 | 0.735±0.01 | 0.706±0.02 | 0.694±0.02 | 0.688±0.02 | 0.716±0.01 | 0.724±0.01 | 0.718±0.01 | 0.693±0.02 | 0.674±0.01 | 0.650±0.01 |
| $\gamma$=0.5 | 0.716±0.01 | 0.738±0.02 | 0.714±0.01 | 0.700±0.01 | 0.695±0.02 | 0.717±0.01 | 0.731±0.01 | 0.726±0.01 | 0.708±0.02 | 0.677±0.03 | 0.657±0.04 |
| $\gamma$=1 | 0.711±0.01 | 0.731±0.01 | 0.692±0.01 | 0.679±0.01 | 0.676±0.02 | 0.721±0.01 | 0.719±0.01 | 0.722±0.01 | 0.697±0.02 | 0.675±0.01 | 0.650±0.02 |

**Memory and I/O.** The layer (1) loads $(A, B, C)$ from HBM to SRAM; (2) streams the per-token coefficients $(\bar{A}, \bar{B})$ of shape $(L, 2d, d_{\text{ssm}})$ through SRAM; (3) performs the selective scan; and (4) writes $(L, 2d)$ output activations back to HBM. The total traffic is therefore

$$\mathcal{O}(L(2d) + 2d\,d_{\text{ssm}}) = \mathcal{O}(Ld + d\,d_{\text{ssm}}) = \mathcal{O}(Ld) \quad (\text{since } d_{\text{ssm}} \ll L).$$

**Time.**

- Computing the discretised coefficients $(\bar{A}, \bar{B}, \bar{C})$ costs

$$\mathcal{O}(3L(2d)\,d_{\text{ssm}}).$$

- The subsequent selective scan costs

$$\mathcal{O}(L(2d)\,d_{\text{ssm}}).$$

Adding the two (and suppressing constant factors) gives the overall layer time complexity

$$\boxed{\mathcal{O}(Ld\,d_{\text{ssm}})}.$$

**Transformer comparison.** A vanilla self-attention layer requires $\mathcal{O}(L^2 d)$ time and memory, so $\Delta t$-Mamba3D—being remains far more scalable for long sequences of imaging data.

## S5 Code Implementation

Figure S1 presents the key code implementation of our proposed $\Delta t$-Mamba3D. The full code will be released upon acceptance of the paper. For both datasets, we followed the preprocessing pipeline provided by the `LIBRA` package.[1]

---

[1] https://www.med.upenn.edu/sbia/libra.html

```python
class StateFusion3D(nn.Module):
    def __init__(self, dim: int,
                 ks=((1,3,3), (3,3,3)),
                 groupwise: bool = True):
        super().__init__()
        groups = dim if groupwise else 1
        convs = []
        for kt, kh, kw in ks:
            pad = (kt//2, kh//2, kw//2)
            convs.append(nn.Conv3d(dim, dim,
                                    kernel_size=(kt,kh,kw),
                                    padding=pad, dilation=1,
                                    groups=groups, bias=False))
        self.convs = nn.ModuleList(convs)
        self.alpha = nn.Parameter(torch.ones(len(convs)))

    def forward(self, h):  # (B,C,T,H,W)
        w = F.softmax(self.alpha, dim=0)
        out = 0
        for wi, conv in zip(w, self.convs):
            out = out + wi * conv(h)
        return out
#  Mamba SSM mix over flattened spatio-temporal tokens             #
# ---------------------------------------------------------------- #
class StructureAwareSSM3D(nn.Module):
    def __init__():
        super().__init__()
        # Skip other parameter
        self.state_fusion3D = (
            StateFusion3D(C_exp, **(fuse_kwargs or {}))
            if fuse else nn.Identity()
        )

    def ssm(self, x: torch.Tensor, dt: torch.Tensor | None = None) -> torch.Tensor:
    # x : (B, C_exp, T, H, W)   — C_exp = 2 * d_model (e.g. 1536)
        B, C_exp, T, H, W = x.shape
        L = T * H * W                   # sequence length
        # 1) Flatten spatio-temporal grid → sequence
        xs = x.view(B, C_exp, L).transpose(1, 2).contiguous()   # (B, L, C_exp)
        x_dbl = self.x_proj(xs)
        # 2) Point-wise projection: (Δt | B | C)
        dts, Bs, Cs = torch.split(
            x_dbl, [self.dt_rank, self.d_state, self.d_state], dim=-1
        )    # (B, L, d_state) each
        dts = self.dt_projs(dts)
        # b) Δt comes from the caller (shape-check later)
        if dt is None:
            raise ValueError("Pass Δt tensor when using continuous-time Mamba")
        # 1) flatten dt to match L  (=T*H*W)
        dt_flat = dt.reshape(x.size(0), -1, 1)            # (B, L, 1)
        # 2) copy along dt_rank
        dt_rep  = dt_flat.repeat(1, 1, self.dt_rank)      # (B, L, dt_rank)
        # 3) linear projection into the channel space
        dtf     = self.dt_projs(dt_rep)
        dtf = dtf * (1 + gamma * dts/12) # main equation
        # 3) Selective scan (kernel from mamba-ssm ≥ 1.1.x)
        h = self.selective_scan(
            xs.transpose(1, 2),         # u  : (B, C_exp, L)
            dtf.transpose(1, 2),        # dt : (B, C_exp, L)
            -torch.exp(self.A_logs),    # A  : (C_exp, d_state)
            Bs.transpose(1, 2),         # B  : (B, d_state, L)
            Cs.transpose(1, 2),         # C  : (B, d_state, L)   -- **must NOT be None**
            self.Ds,                    # D  : (C_exp)
            z=None,
            delta_bias=self.dt_projs.bias,
            delta_softplus=True,
            return_last_state=False,
        )                               # (B, C_exp, L)

        # 4) Gate projection   (16 → 1536)
        C_gate = self.C_proj(Cs).transpose(1, 2)        # (B, C_exp, L)
        # 5) Restore 5-D layout
        h      = h     .reshape(B, C_exp, T, H, W)
        C_gate = C_gate.reshape(B, C_exp, T, H, W)
        # 6) Optional multi-scale fusion + residual
        h = self.state_fusion3D(h)
        y = h * C_gate + x * self.Ds.view(1, -1, 1, 1, 1)
        return y
```

Figure S1: Key code for proposed method

5


\end{document}